%% file: 00main.tex
\documentclass{article} % For LaTeX2e
\usepackage{iclr2027_conference,times}
\iclrfinalcopy

\input{math_commands.tex}

\usepackage[colorlinks,
            linkcolor=cyan,
            anchorcolor=blue,
            citecolor=magenta
            ]{hyperref}       % hyperlinks
\usepackage{url}
\usepackage{graphicx}
\usepackage{subcaption}
\usepackage{booktabs}
\usepackage[table]{xcolor}
\usepackage{multirow}
\usepackage{wrapfig}
\usepackage{titletoc}

\newcommand{\nm}{{GSCBench}}

\title{Revisiting the Generalization of Neural Graph Edit Distance Models}

\author{
Zhouyang Liu$^{1}$,
Ning Liu$^{2}$,
Yixin Chen$^{1}$,
Jiezhong He$^{1}$,
Dongsheng Li$^{1}$ \\
$^{1}$National University of Defense Technology \\
$^{2}$Information Support Force Engineering University \\
\texttt{\{liuzhouyang20,liuning17a,chenyixin,jiezhonghe,dsli\}@nudt.edu.cn}
}
\begin{document}

\maketitle
\lhead{Under Review.}
\begin{abstract}
Neural approaches to Graph Edit Distance (GED) have achieved strong results under standard within-dataset evaluation, but much less is known about how well these models transfer across graph collections. We conduct a systematic study of this problem using exact GED supervision across diverse graph datasets and a broad set of representative learning-based methods. Our results reveal a pronounced gap between within-collection performance and cross-collection transfer. Models that perform well on their training collections often lose this advantage when evaluated on structurally different data. Training on multiple source collections substantially improves zero-shot transfer and provides a better starting point when limited supervision is available for a new target collection. Further analysis shows that transfer behavior varies with the source--target direction and the structural characteristics of the collections involved. These findings suggest that conventional within-collection evaluation provides only a partial view of the generalization behavior of neural GED models and motivate broader evaluation across heterogeneous graph collections. Code and benchmark data are publicly available at \href{https://github.com/liuzhouyang/GSCBench}{GitHub} and \href{https://huggingface.co/datasets/ush3r/GSCbench}{Hugging Face}.
\end{abstract}

\section{Introduction}
Graph Edit Distance (GED) is the minimum number of node and edge edits required to transform one graph into another, providing an interpretable measure of structural dissimilarity that is widely used in graph matching \citep{hap}, similarity search \citep{nc-iso}. GED is domain-agnostic because it does not require interpreting domain-specific label semantics, allowing the same edit-based measure to be applied across graph collections and domains. Computing GED exactly is NP-hard \citep{nphard}, motivating neural methods that accelerate GED computation by learning from previously solved graph pairs \citep{simgnn,gmn,noah,liu2025graph}. Yet the graph pairs used for training inevitably reflect collection-specific graph sizes, labels, structural patterns, and distance distributions, which can vary across collections. In mainstream graph learning, such shifts often lead to substantial performance drops \citep{when_to_pretrain_gnn,ju2025graphbridge}. For GED, the task is unchanged, while structural differences are precisely what the distance is supposed to capture. Does what the model learns travel with GED, or stop at the dataset boundary?

Standard evaluation protocols offer little evidence on this question. Learning-based GED methods use neural models in different ways to estimate or compute GED. Some predict the distance directly \citep{simgnn,gmn,ranjan2022greed,liu2026rethinkingflexiblegraphsimilarity}, while others learn signals for constructing or refining node assignments \citep{gotsim,gedgnn,gediot,gedranker}. Search-based methods retain explicit combinatorial search and use learned information to guide the search process \citep{noah,genn,mata,pellizzoni2026gelato}. Despite these differences, evaluation usually keeps training and test pairs within the same graph collection. Such evaluation establishes how well a method handles unseen pairs from the same collection, but not whether its performance persists when the collection changes. Recent studies have evaluated generalization under changing edit-cost settings \citep{liu2026rethinkingflexiblegraphsimilarity} and across graph collections \citep{verma2025grail,pellizzoni2026gelato}. Yet collection shift has so far been studied mainly as an evaluation setting for individual methods, leaving its broader effect on neural GED learning unclear.

A systematic study of collection shift requires changing the graph collection while keeping benchmark construction and supervision comparable. Existing GED resources make this difficult for two reasons. (1) \emph{Limited coverage and inconsistent construction.} The graph collections introduced with SimGNN \citep{simgnn} remain the standard testbed for many neural GED methods, yet their exact annotations cover only graphs with at most 10 nodes, with larger graphs relying on approximate GED labels. The presence of isomorphic graphs further allows structurally identical instances to appear across training and test splits, leading to data leakage \citep{pmlr-v267-roy25a}. Subsequent resources broaden exact GED coverage, but were constructed independently with different graph filtering and annotation procedures. As a result, collection and resource construction change together when these datasets are combined for cross-collection evaluation. (2) \emph{Method-dependent supervision.} Scalar GED labels are sufficient for some neural methods, while others require additional supervision, such as edit counts or ground-truth node mappings \citep{tagsim,gedgnn}. These signals are not consistently available across existing resources, preventing the same set of collections from being used for a broad cross-method comparison.

Here, we introduce an open benchmark for studying cross-collection generalization in neural GED computation. It provides exact GED annotations for 11 graph collections, comprising 5,523 graphs and 1,903,452 graph pairs, with graphs containing up to 71 nodes. For each pair, we also retain the corresponding node alignment, enabling methods with different supervision requirements to be evaluated on the same collections. We use this benchmark to study three questions:
\begin{itemize}
\item \emph{Single-source generalization.} When exact GED supervision comes from a single graph collection, how well does the learned model transfer to other collections, and do methods that perform well in-collection remain strong after transfer?
\item \emph{Multi-source generalization.} Can exact GED supervision from multiple graph collections be learned jointly, and how does this affect performance on both source collections and unseen target collections?
\item \emph{Limited-target adaptation.} When only a small amount of exact GED supervision is available for a new target collection, how much does knowledge learned from other collections help compared with learning from the target data alone?
\end{itemize}
Our evaluation reveals a sharp gap between the collection-invariant definition of GED and the behavior of models learned from exact GED supervision. Among the 20 evaluated methods, 18 outperform the best approximation results on their source collections, yet none retains this advantage after transfer. This picture changes markedly once the same methods are trained on multiple collections. Almost every method improves on unseen targets, even without introducing a dedicated mechanism for cross-collection generalization. The resulting knowledge also remains useful after target supervision becomes available, reducing median MAE by 42.9\% with only 100 target pairs. 

Our contributions are threefold. First, we provide a cross-collection GED benchmark with exact distances and node alignments across 11 graph collections. Second, we release an open-source framework that integrates representative neural GED methods under a unified training and evaluation pipeline, with support for single-source transfer, multi-source training, and limited-target adaptation. Third, we provide the first systematic empirical study of neural GED generalization across single-source transfer, multi-source training, and limited-target adaptation.
\section{Background}

\paragraph{GED Formulation} Given a query graph $Q=(\mathcal{V}_Q, \mathcal{E}_Q)$ and a data graph $D=(\mathcal{V}_D, \mathcal{E}_D)$, Graph Edit Distance (GED) measures the minimum number of edit operations required to transform $Q$ into a graph isomorphic to $D$. We consider undirected graphs with categorical node labels and unlabeled edges. The allowed edit operations include node insertions, deletions, and substitutions, as well as edge insertions and deletions. Let $\mathcal{F}(Q,D)$ denote the set of valid edit paths from $Q$ to $D$. GED is then defined as
\begin{equation}
\mathrm{GED}(Q,D)
=
\min_{f\in\mathcal{F}(Q,D)} |f|,
\end{equation}
where $|f|$ denotes the number of edit operations in $f$. An optimal edit path also induces a node alignment between $Q$ and $D$.

\paragraph{Learning-based GED computation} 
Given a query graph collection $\mathcal{Q}$ and a data graph collection $\mathcal{D}$, learning-based GED computation aims to estimate $\mathrm{GED}(Q,D)$ for graph pairs $Q\in\mathcal{Q}$ and $D\in\mathcal{D}$ using learned models to replace or assist conventional GED computation. The learned signal can capture different levels of information about a graph pair. It may remain largely at the graph level \citep{gmn,ranjan2022greed}, explicitly model cross-graph interactions \citep{simgnn,gmn,h2mn}, or incorporate broader structural context and candidate-level matching information \citep{graphsim,mata,g2r,liu2026rethinkingflexiblegraphsimilarity,pellizzoni2026gelato}. How this information is used also varies across methods. Some models directly estimate GED, while others construct or evaluate explicit node correspondences and complete matching candidates \citep{gotsim,graphedx,gedgnn,gediot,gedranker}; search-based methods instead use learned signals to guide the exploration of candidate solutions or partial states \citep{noah,genn,mata,pellizzoni2026gelato}. These differences lead to different inference procedures and supervision requirements.

\begin{wrapfigure}{r}{0.5\textwidth}
\centering
\includegraphics[width=\linewidth]{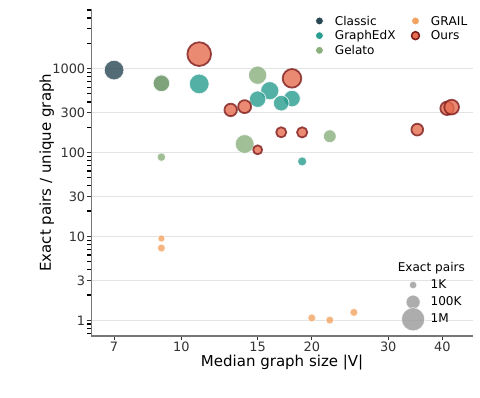}
\caption{Landscape of existing exact GED resources.}
\label{fig:landscape}
\end{wrapfigure}

\section{\nm}

\subsection{Existing GED Resources}

\paragraph{Limited collection coverage and consistency}
Existing GED resources have gradually expanded the scale of exact supervision. The benchmark introduced with SimGNN \citep{simgnn} covers three domains, but exact GED is available only for graphs containing at most 10 nodes. \cite{pmlr-v267-roy25a} remove the isomorphic duplicates responsible for train--test leakage and extend exact annotations to graphs of about 20 nodes. Gelato provides optimal solutions for graphs up to 30 nodes \citep{pellizzoni2026gelato}, while GRAIL incorporates collections with average graph sizes reaching 37 nodes but annotates only sampled graph pairs \citep{verma2025grail}. Our benchmark reaches 71 nodes with substantially denser exact-pair coverage, as summarized in Figure~\ref{fig:landscape}. Beyond coverage, these resources also follow different construction procedures. Gelato intentionally keeps isomorphic graphs within training splits, while several existing resources include disconnected graphs. Exact GED is also generated using different solvers, including A\* and MIP-F2, with different or unspecified timeout budgets. These choices alter which graphs and graph pairs receive exact annotations, introducing construction differences into cross-collection comparisons.

\paragraph{Method-dependent supervision}
Learning-based GED methods require different forms of supervision. Direct prediction methods can be trained with scalar GED labels, while other methods additionally use edit-level information \citep{tagsim} or node correspondences \citep{gedgnn}. Existing resources do not provide these signals consistently across collections. Most primarily provide scalar GED labels, while node mappings are available only in resources such as GEDGNN and GELATO for their respective processed collections \citep{gedgnn,pellizzoni2026gelato}. Consequently, the graph collections available for evaluation can depend on the supervision required by each method, preventing a common cross-method comparison. Our benchmark provides exact GED and node alignments for every annotated pair across all 11 collections, supporting different supervision requirements on the same data.

\subsection{Benchmark Construction}
\input{stats}

\begin{figure}[h]
    \centering
    \includegraphics[width=\linewidth]{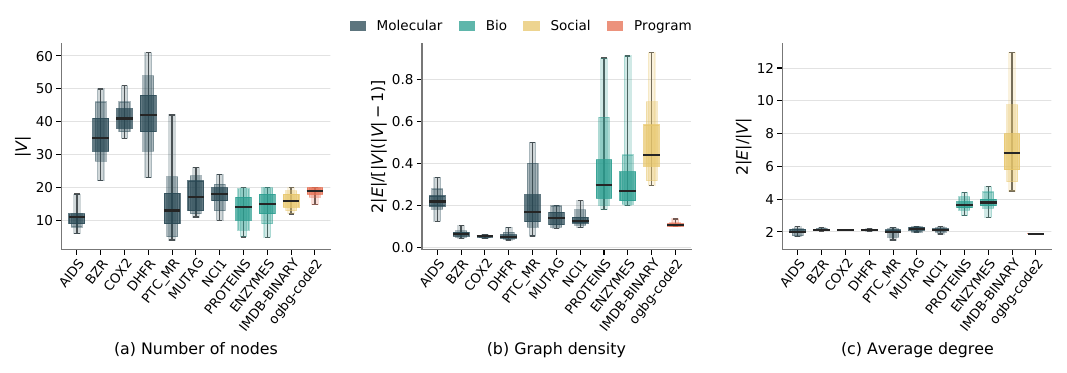}
    \caption{Graph-level characteristics across the 11 collections.} \label{fig:graph_characteristics}
\end{figure}

\paragraph{Graph Collections}
We collect candidate graphs from 11 collections in TUDataset \citep{tudataset} and the Open Graph Benchmark (OGB) \citep{ogbdataset}, spanning molecular, biological, social, and program graphs. We remove disconnected graphs and isomorphic duplicates before forming graph pairs. Table~\ref{tab:dataset_statistics} reports the statistics of the collections and their exact GED annotations, while Fig.~\ref{fig:graph_characteristics} presents their graph-level characteristics.

\paragraph{Exact GED Annotations}
We form graph pairs within each retained collection and compute their exact GED using AStar-LSa \citep{astarlsa}, with a 300-second time limit per pair. Only pairs solved exactly within this limit are included in the benchmark. This procedure yields 5,523 graphs and 1,903,452 exact graph pairs across the 11 collections. For each pair, we store both its GED value and the corresponding node alignment, supporting methods trained from graph-level distances as well as methods that learn node correspondences or construct edit paths.

\subsection{Data Analysis}\label{sec:analysis}

Beyond the basic statistics in Table \ref{tab:dataset_statistics} and Figure ~\ref{fig:graph_characteristics}, we further examine how graph structure relates to GED and how these structural characteristics compare across collections.

\paragraph{Structural Factors Associated with GED}
We first examine which properties of a graph pair are associated with GED in each collection. To this end, we consider six factors ranging from coarse count statistics to local and global topology: mean $|V|$ and mean $|E|$ characterize the \emph{absolute scale} of the pair, $|\Delta V|$ and $|\Delta E|$ measure \emph{node- and edge-count difference}, and graphlet and spectral distances capture differences in \emph{local structural} composition and \emph{global connectivity}, respectively. Detailed description can be found in Appendix.

\begin{figure}[h]
    \centering
    \includegraphics[width=\linewidth]{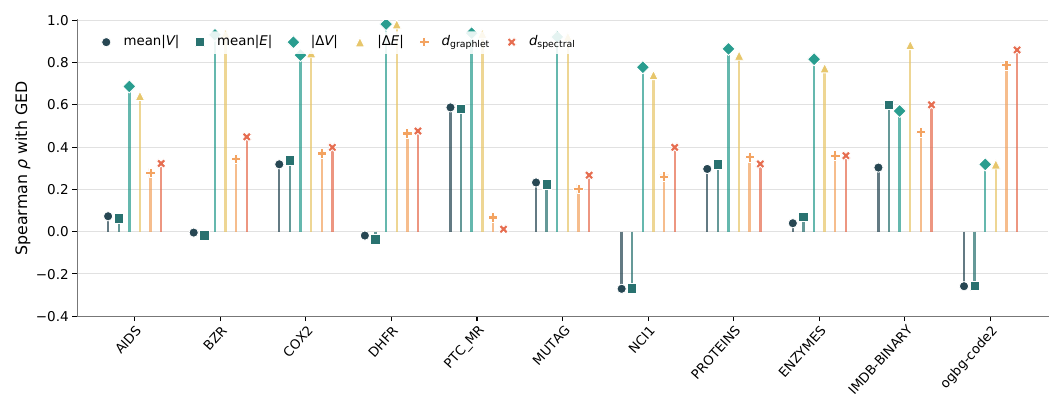} 
    \caption{Spearman correlations between GED and six pair-level structural signals.} \label{fig:pair_correlation}
\end{figure}

Figure~\ref{fig:pair_correlation} reports the resulting Spearman correlations for each graphlet type and collection. The dominant graphlet patterns differ markedly across collections. In code2, the four-node path and three-star both reach a correlation of $0.786$. In IMDB-Binary, the four-node path is nearly uncorrelated with GED at $0.010$, while the four-node clique, triangle, and three-node path reach $0.471$, $0.410$, and $0.410$, respectively. These contrasts show that the local structural patterns most associated with GED vary across collections, consistent with the collection-dependent behavior of the aggregate graphlet distance.

More importantly, the strength of these structural signals varies substantially across collections. In PTC-MR, $|\Delta V|$ and $|\Delta E|$ correlate with GED at $0.938$ and $0.936$, while the graphlet and spectral correlations are only $0.067$ and $0.012$. ogbg-code2 shows the opposite pattern. Its count differences correlate with GED at only $0.318$, while the graphlet and spectral correlations reach $0.786$ and $0.859$. This contrast shows that the structural signals associated with GED can differ markedly across collections. Generalization may therefore depend on which structural signals a model relies on when estimating GED.

\begin{figure}[h]
    \centering
    \includegraphics[width=\linewidth]{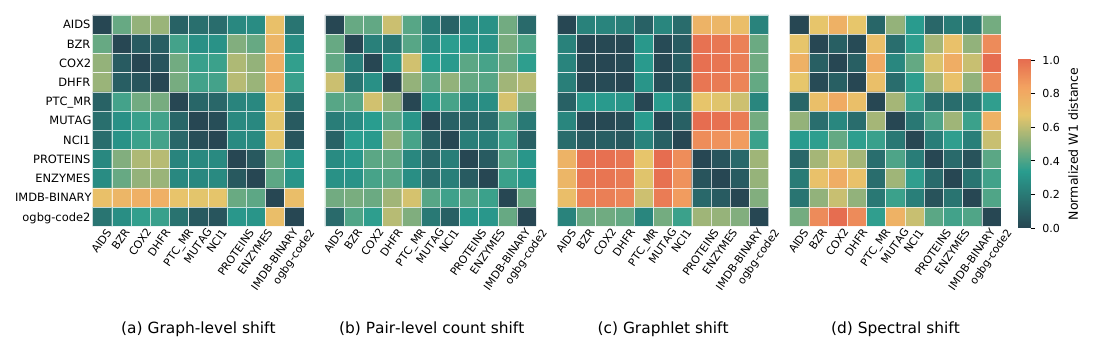}
    \caption{Relative distribution shifts across graph collections under four structural views (the darker the more similar).}\label{fig:collection_shift}
\end{figure}

\paragraph{Collection Similarity under Different Structural Views}
We next turn from within-collection relationships to structural distribution shifts across collections. We compare collection similarity from four views. The \emph{graph-level view} considers the distributions of $|V|$, $|E|$, density, and average degree over individual graphs, while the \emph{pair-level count view} considers mean $|V|$, mean $|E|$, $|\Delta V|$, $|\Delta E|$, and the node-count ratio between the larger and smaller graph in each exact pair. The other two views compare the distributions of \emph{graphlet and spectral} distances over exact graph pairs, capturing local and global structural differences, respectively. For each descriptor, we use the Wasserstein-1 distance to measure the shift between its empirical distributions in two collections. We normalize each descriptor by its maximum pairwise distance across the 11 collections, allowing different descriptors to be compared and combined on a common relative scale. Details of the computation are provided in the Appendix.

Figure~\ref{fig:collection_shift} shows that collection similarity depends strongly on the structural view. At the graph level, the average relative shift is $0.266$ for collection pairs from the same domain, compared with $0.418$ for cross-domain pairs. This difference almost disappears at the pair-count level, where the corresponding means are $0.342$ and $0.347$. Across the 55 collection pairs, the relative-shift rankings under these two views have a Spearman correlation of $\rho=0.757$, although individual relationships can still change substantially. For example, COX2 and DHFR are highly similar at the graph level yet much farther apart at the pair-count level. The divergence is stronger for higher-order structure, where the rankings under graphlet and spectral shifts have a Spearman correlation of only $\rho=0.401$. Hence, two collections that appear close under one structural view need not remain close under another, and no single notion of collection similarity captures all source--target shifts.

\subsection{License and Availability}
The graph collections in our benchmark are sourced from TUDataset \citep{tudataset} and OGB \citep{ogbdataset} and remain subject to the licenses and terms of their original sources. We release the generated GED annotations and benchmark metadata under the CC BY-NC 4.0 License, and the training and evaluation framework under the MIT License. Third-party methods remain subject to their respective original licenses.

\section{Experiments}
\subsection{Experimental Setup}
\paragraph{Data Splits}
For each collection, 20\% of the graphs are held out from training, of which 20\% are used for validation and the remainder for testing. Graph pairs are constructed only after the graph-level split, ensuring that no test graph appears in the training pairs. We evaluate on all exact pairs containing at least one test graph. The appendix further separates these pairs into pairs between training or validation graphs and test graphs ($D$--$Q$), and pairs between two test graphs ($Q$--$Q$). Each experiment is repeated over five runs to account for both training randomness and graph-split variation. Specifically, three runs vary the training seed under a fixed split, while three vary the graph split under a fixed training seed, with one run shared between the two groups.

\paragraph{Baselines}
We benchmark 21 representative learning-based GED methods spanning three major computational paradigms. \emph{Direct prediction methods} estimate GED directly from learned graph-pair representations, including SimGNN \citep{simgnn}, GMN-emb \citep{gmn}, GMN-match \citep{gmn}, GraphSim \citep{graphsim}, H2MN \citep{h2mn}, TaGSim \citep{tagsim},  EGSC \citep{egsc}, ERIC \citep{eric}, GREED \citep{ranjan2022greed},  Graph2Region \citep{g2r}, GraSP \citep{grasp}, and GEN \citep{liu2026rethinkingflexiblegraphsimilarity}. \emph{Matching-based methods} construct node correspondences or edit solutions from learned matching signals, including GOTSim \citep{gotsim}, GEDGNN \citep{gedgnn}, GraphEdX \citep{graphedx} and GEDIOT \citep{gediot}. \emph{Search-based methods} use learned signals to guide an explicit combinatorial search, including NOAH \citep{noah}, GENN-A* \citep{genn}, MATA* \citep{mata}, and Gelato \citep{pellizzoni2026gelato}. 

\paragraph{Implementation Details} \emph{For categorical node labels, we use one-hot node features whose dimensionality is determined by the maximum node-label value across all benchmark collections.} We use the official implementations released by the original authors and integrate them into a unified benchmark, with necessary optimization for methods that do not natively support batched training. The benchmark implementation will be publicly released. We follow the hyperparameters recommended by the original papers and repositories. We use a training batch size of 128, an evaluation batch size of 2048, and a learning rate of $10^{-3}$; batch sizes are reduced when necessary to avoid out-of-memory errors. We apply early stopping based on validation MSE, with a patience of 50 epochs for standard training and 20 epochs for pretraining. All experiments are conducted on a server with three NVIDIA A100-SXM4-40GB GPUs and an Intel Xeon Gold 6348 CPU.

\paragraph{Evaluation Metrics}
Our primary metrics are Mean Absolute Error (MAE) for GED estimation and Precision$@$10 ($P@10$) for similarity retrieval. MAE measures the absolute deviation between predicted and exact GED values, while $P@10$ measures the fraction of the exact top-10 nearest graphs recovered in the predicted top-10. We report metric values using a split-balanced median across the five runs. For ranking analyses, methods are ranked separately within each run and evaluation task, and the resulting ranks are averaged across runs and tasks to summarize overall performance. Details of the aggregation procedure are provided in Appendix~\ref{app:aggregation}.

\input{single/single_source_rank_table}
\begin{wrapfigure}{r}{0.485\textwidth}
\centering
\includegraphics[width=\linewidth]{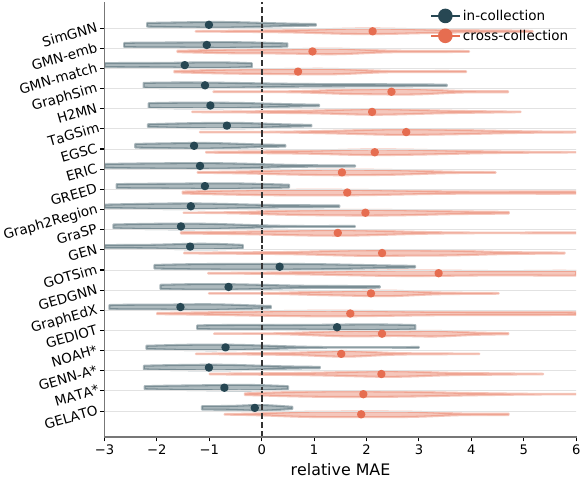}
\caption{In- and Cross-collection MAE to approximation.}\label{fig:in_and_cross}
\end{wrapfigure}
\subsection{Single-Source Generalization}
We evaluate how well a GED model transfers from one graph collection to another. Each method is trained on one of the 11 collections and evaluated on all 11, yielding 121 directed source--target tasks. Figure~\ref{fig:in_and_cross} compares neural MAE with the best approximation result on each target collection. In the in-collection setting, 18 of the 20 evaluated neural methods achieve a lower median MAE than this target-specific cutoff. After cross-collection transfer, none does. Even GMN-match and GMN-emb, which achieve the lowest cross-collection median MAE among the neural methods, reach approximately $1.62\times$ and $1.96\times$ the approximation MAE. These results show that the advantage of neural GED estimation is largely confined to the collection on which the model is trained.

The impact of collection shift also varies substantially across methods.  Table~\ref{tab:single_source_rank}, compares their relative performance before and after transfer. GEN has an average MAE rank of 3.16 in-collection and reaches the top five on 10 of the 11 collections, but its cross-collection rank drops to 12.67 and its Top-5 Rate to 10.0\%. GMN-emb exhibits the opposite trend, improving from an average MAE rank of 10.20 to 4.02 and from a Top-5 Rate of 9.1\% to 75.5\%. Thus, collection shift not only reduces absolute performance, but also changes which methods remain competitive after transfer. Strong in-collection performance is therefore not a reliable indicator of cross-collection generalization. Additional results are provided in Appendix.

\begin{figure}[h]
    \centering
    \begin{subfigure}[t]{0.485\columnwidth}
        \centering
        \includegraphics[width=\linewidth]{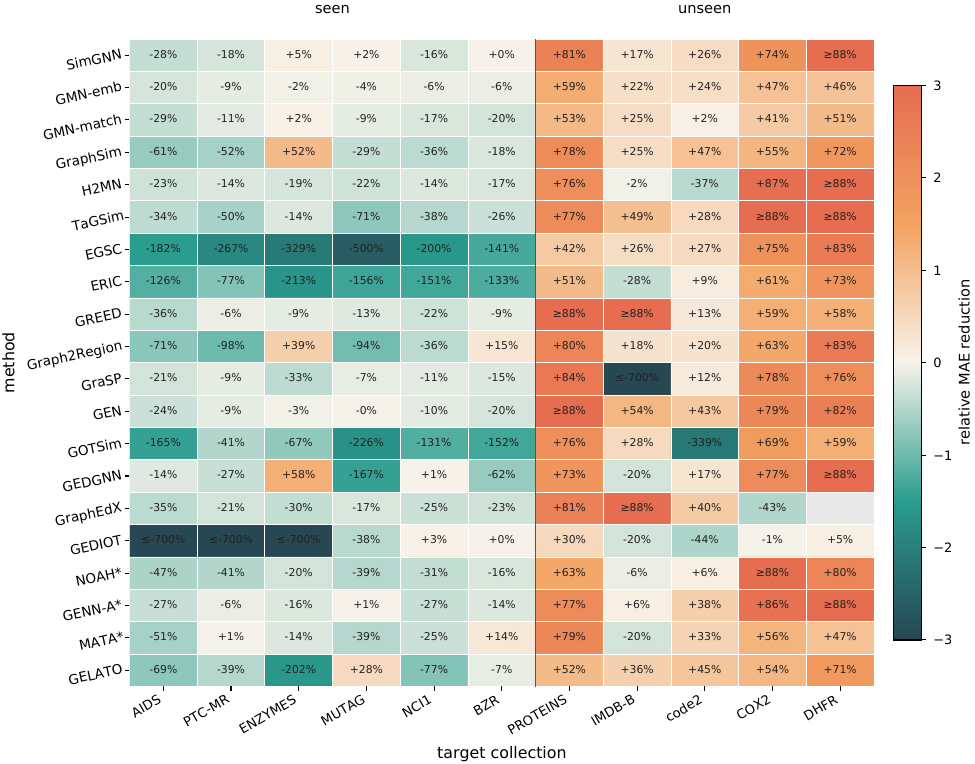}
        \caption{Relative MAE Reduction.}
        \label{fig:mae_gain}
    \end{subfigure}
    \hfill
    \begin{subfigure}[t]{0.485\columnwidth}
        \centering
        \includegraphics[width=\linewidth]{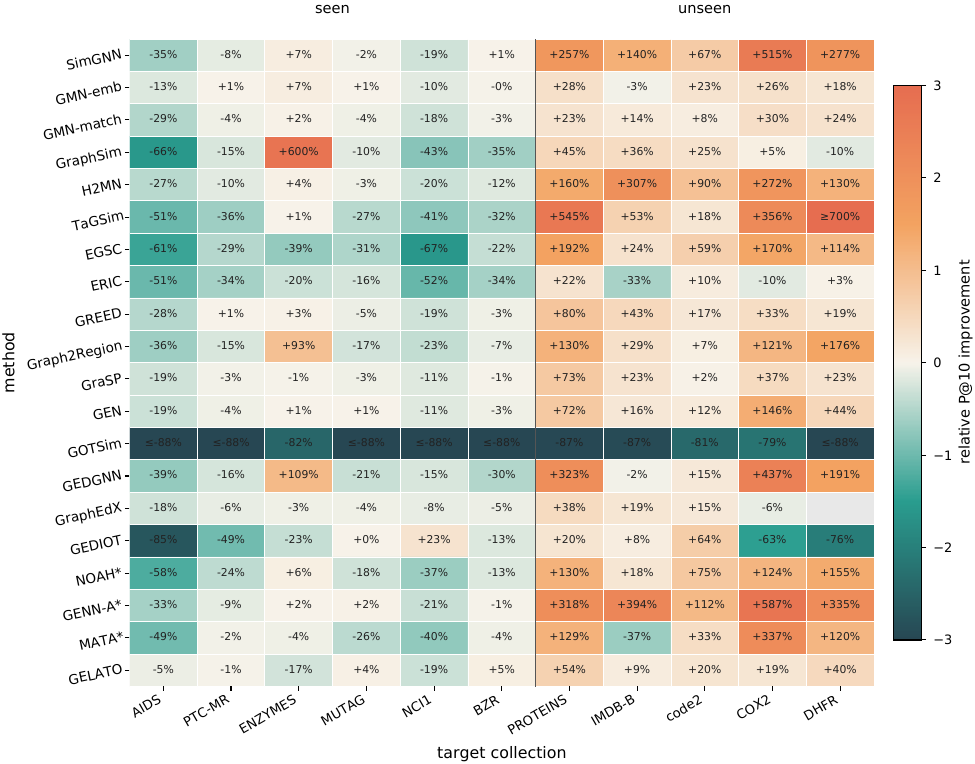}
        \caption{Relative $P@10$ Improvement.}
        \label{fig:rank_gain}
    \end{subfigure}

    \caption{
    Multi-source training relative to single-source baselines.
    }
    \label{fig:multisource_gain}
\end{figure}
\input{multi/multisource_rank_summary}
\subsection{Multi-Source Generalization}
We next examine whether learning from multiple graph collections improves generalization beyond a single source. We pretrain each method on a fixed source pool consisting of AIDS, BZR, ENZYMES, MUTAG, NCI1, PTC-MR, and LINUX. LINUX is included only as an additional pretraining source and is not part of the 11 benchmark collections used for evaluation. The remaining six source collections therefore serve as seen targets, while PROTEINS, IMDB-Binary, ogbg-code2, COX2, and DHFR are held out as unseen targets. For seen collections, we compare multi-source training with target-specific single-source training.

Figure~\ref{fig:multisource_gain} reveals a clear difference between the two settings. On seen collections, multi-source training rarely improves over target-specific training in either MAE or $P@10$. The pattern reverses on unseen collections, where multi-source training improves both metrics for the large majority of method--target evaluations. Thus, combining supervision from multiple collections sacrifices some collection-specific specialization, but substantially improves transfer to collections that are absent from training. The gains on unseen collections also vary across methods. Table~\ref{tab:multisource_rank_summary} shows that GEN improves its average unseen MAE rank from 14.40 to 8.80, while GENN-A* moves from 13.00 to 9.80. In contrast, GMN-emb and GMN-match move from 2.00 to 3.40 and from 1.80 to 3.80, respectively. Multi-source training therefore improves unseen-target transfer broadly, but redistributes the relative advantages among methods. Detailed implementation and additional results are provided in Appendix.

\subsection{limited-target Adaptation}

\begin{figure}[h]
    \centering
    \includegraphics[width=\linewidth]{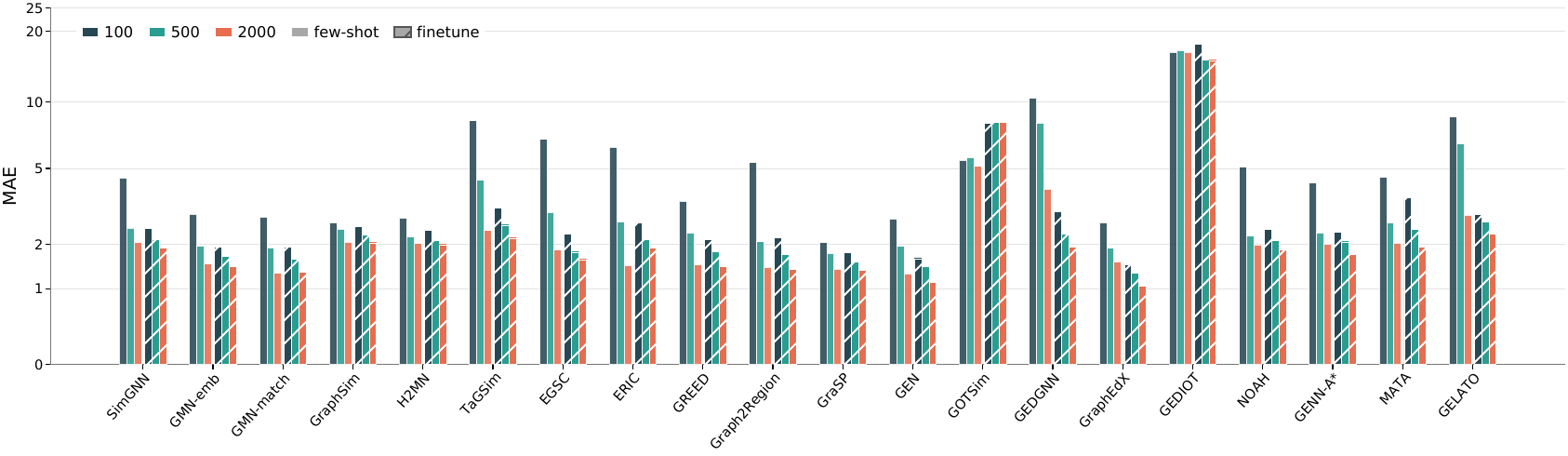} 
    \caption{Limited-target adaptation with and without multi-source pretraining.} \label{fig:limited_mae}
    \vspace{-0.3cm}
\end{figure}

We finally examine whether knowledge learned from multiple source collections remains useful when limited supervision from an unseen target collection becomes available. For each of the five unseen targets, we compare target-only training from random initialization with fine-tuning from the multi-source pretrained model using the same 100, 500, or 2,000 exact target graph pairs.

Figure~\ref{fig:limited_mae} shows that multi-source pretraining provides the largest benefit when target supervision is scarce. With only 100 target pairs, fine-tuning reduces the median MAE across target collections for 18 of the 20 methods, with a median relative reduction of 42.9\%. The advantage remains widespread at 500 pairs, where 19 methods benefit, but the median reduction decreases to 14.4\%. At 2,000 pairs, the gap narrows further to 6.2\%, although fine-tuning still improves 17 methods. Multi-source pretraining therefore substantially improves sample efficiency under limited target supervision, while its advantage diminishes as more target-specific GED labels become available. Additional per-target results are provided in Appendix.

\section{Future Opportunities \& Limitations}

To enable systematic study of cross-collection generalization in neural GED computation, we introduce a benchmark covering 11 graph collections, 5,523 graphs, and 1,903,452 exact graph pairs, together with a unified evaluation framework for representative learning-based methods. Our experiments show that current neural GED methods generalize poorly across collections, while broader multi-source supervision can substantially improve transfer to unseen collections. Based on these observations and our data analysis, we highlight several directions for future research.

\begin{itemize}

\item \textbf{What structural information should GED representations preserve?}
Representation learning remains central across direct prediction, explicit matching, search-based methods, and unsupervised methods. Our analysis shows that richer structural descriptors are not necessarily more informative for GED, as graphlet and spectral distances have weak associations with GED in many collections and their importance varies across collections. Identifying which structural signals consistently support GED computation, and how representations should preserve them across collection shifts, is a fundamental challenge.

\item \textbf{Can collection characteristics predict and improve transferability?}
Our analysis shows that  collection characteristics change substantially across graph-level, pair-level, and higher-order structural views. A key question is which structural characteristics best predict source--target transfer. Such signals could also guide test-time adaptation, allowing pretrained GED models to adjust to the structure of a new collection without exact target GED supervision.

\end{itemize}

Our study focuses on graph pairs solved exactly within the computation budget and on undirected graphs with categorical node labels, unlabeled edges, and unit edit costs. The multi-source and adaptation experiments also use fixed source--target configurations. Extending the benchmark to larger graphs, richer graph settings, and broader source combinations remains future work.

\section*{AI use statement}

Generative AI tools were used to improve the clarity and readability of the manuscript, including language polishing and sentence-level rewriting, and to assist with the organization and presentation of experimental results. They were also used to suggest the structure and title of the paper and to create or refine scientific figures, including their visual layout and presentation style. The research questions, hypotheses, benchmark design, experimental methodology, underlying experimental results, and scientific conclusions were determined and verified by the authors. All AI-assisted text, structural suggestions, and figures were reviewed and revised by the authors. The authors take full responsibility for the final content of the paper.

\section*{Reproducibility Statement}

We provide detailed descriptions of the benchmark construction, including graph filtering, pair construction, exact GED computation, and data splits. The experimental setup specifies the training procedure, evaluation metrics, and protocols for single-source generalization, multi-source generalization, and limited-target adaptation. Additional implementation details and per-collection results are included in the Appendix.

\bibliography{references}
\bibliographystyle{iclr2027_conference}
\newpage
\appendix

\input{10appendix}

\end{document}

%% file: math_commands.tex
\usepackage{amsmath,amsfonts,bm}

\def\eqref#1{equation~\ref{#1}}
\def\1{\bm{1}}

\DeclareMathAlphabet{\mathsfit}{\encodingdefault}{\sfdefault}{m}{sl}
\SetMathAlphabet{\mathsfit}{bold}{\encodingdefault}{\sfdefault}{bx}{n}

%% file: stats.tex
\begin{table*}[t]
\centering
\caption{Statistics of the graph collections in \nm. Graph size and GED are reported as median / 90th percentile / maximum.}
\label{tab:dataset_statistics}
\resizebox{\linewidth}{!}{ \renewcommand{\arraystretch}{1.0}
\begin{tabular}{llrcccc}
\toprule
Dataset & Domain & $\#$Graphs & $\#$Node Labels &
$|\mathcal{V}|$ &
GED &
\#Exact Pairs \\
\midrule
AIDS        & Molecular  & 1,488 & 27        & 11 / 12 / 20   & 10 / 16 / 42  & 1,106,325 \\
BZR         & Molecular  &   404 & 10        & 35 / 46 / 57   & 20 / 45 / 93  &    37,781 \\
COX2        & Molecular  &   465 & 8         & 41 / 46 / 56   & 16 / 27 / 59  &    78,100 \\
DHFR        & Molecular  &   719 & 9         & 42 / 54 / 71   & 28 / 55 / 107 &   124,714 \\
PTC\_MR     & Molecular  &   328 & 18        & 13 / 23.3 / 64 & 19 / 41 / 134 &    52,706 \\
MUTAG       & Molecular  &   175 & 7         & 17 / 23.6 / 28 & 15 / 27 / 43  &    15,220 \\
NCI1        & Molecular  &   992 & 16        & 18 / 21 / 25   & 15 / 22 / 42  &   379,032 \\
PROTEINS    & Biological &   364 & 3         & 14 / 19.7 / 20 & 24 / 37 / 56  &    64,010 \\
ENZYMES     & Biological &   117 & 3         & 15 / 20 / 20   & 26 / 38 / 66  &     6,286 \\
IMDB-BINARY & Social     &   280 & Unlabeled & 16 / 19 / 20   & 27 / 61 / 172 &    22,703 \\
ogbg-code2  & Program    &   191 & 32        & 19 / 20 / 20   & 15 / 27 / 38  &    16,575 \\
\midrule
\textbf{Total}
            & --         & \textbf{5,523} & -- & -- & -- &
              \textbf{1,903,452} \\
\bottomrule
\end{tabular}}
\end{table*}

%% file: single/single_source_rank_table.tex
\begin{table*}[t]
\centering
\caption{Single-Source Generalization method-level rank summary. }
\label{tab:single_source_rank}
\resizebox{\linewidth}{!}{ \renewcommand{\arraystretch}{1.0}
\begin{tabular}{l|ccc|ccc|c}
\toprule
\multirow{2}{*}{Method} & \multicolumn{3}{c|}{In-collection} & \multicolumn{3}{c|}{Cross-collection} & \multirow{2}{*}{$\Delta$ Rank} \\
& MAE Rank & P@10 Rank & Top-5 Rate & MAE Rank & P@10 Rank & Top-5 Rate & \\
\midrule
SimGNN & 12.82 & 14.54 & 0/11 (0.0\%) & 11.51 & 14.56 & 0/110 (0.0\%) & 1.30 \\
GMN-emb & 10.20 & 8.66 & 1/11 (9.1\%) & 4.02 & 3.98 & 83/110 (75.5\%) & 6.18 \\
GMN-match & 4.00 & 6.01 & 7/11 (63.6\%) & 2.96 & 4.33 & 98/110 (89.1\%) & 1.04 \\
GraphSim & 13.22 & 13.63 & 0/11 (0.0\%) & 12.27 & 10.93 & 2/110 (1.8\%) & 0.95 \\
H2MN & 12.31 & 12.41 & 0/11 (0.0\%) & 11.01 & 13.40 & 4/110 (3.6\%) & 1.30 \\
TaGSim & 13.95 & 16.28 & 0/11 (0.0\%) & 13.82 & 17.89 & 2/110 (1.8\%) & 0.12 \\
EGSC & 7.82 & 8.01 & 2/11 (18.2\%) & 11.05 & 13.31 & 3/110 (2.7\%) & -3.23 \\
ERIC & 6.85 & 5.65 & 5/11 (45.5\%) & 8.14 & 5.71 & 23/110 (20.9\%) & -1.29 \\
GREED & 9.82 & 9.79 & 1/11 (9.1\%) & 9.90 & 6.45 & 32/110 (29.1\%) & -0.08 \\
Graph2Region & 6.69 & 7.01 & 6/11 (54.5\%) & 10.29 & 8.90 & 19/110 (17.3\%) & -3.60 \\
GraSP & 5.02 & 6.03 & 7/11 (63.6\%) & 7.92 & 5.55 & 45/110 (40.9\%) & -2.91 \\
GEN & 3.16 & 3.37 & 10/11 (90.9\%) & 12.67 & 7.53 & 11/110 (10.0\%) & -9.51 \\
GOTSim & 17.76 & 16.80 & 0/11 (0.0\%) & 15.00 & 10.90 & 0/110 (0.0\%) & 2.76 \\
GEDGNN & 16.22 & 15.65 & 0/11 (0.0\%) & 11.20 & 14.88 & 0/110 (0.0\%) & 5.02 \\
GraphEdX & 3.51 & 3.99 & 10/11 (90.9\%) & 9.02 & 3.64 & 23/65 (35.4\%) & -5.51 \\
GEDIOT & 13.95 & 13.71 & 0/11 (0.0\%) & 13.33 & 11.51 & 3/110 (2.7\%) & 0.61 \\
NOAH$^*$ & 13.09 & 15.96 & 0/11 (0.0\%) & 7.25 & 13.07 & 33/110 (30.0\%) & 5.85 \\
GENN-A$^*$ & 11.82 & 13.61 & 0/11 (0.0\%) & 11.64 & 14.96 & 1/110 (0.9\%) & 0.17 \\
MATA$^*$ & 13.25 & 12.38 & 0/11 (0.0\%) & 11.52 & 16.12 & 3/110 (2.7\%) & 1.73 \\
GELATO & 14.55 & 6.51 & 1/11 (9.1\%) & 10.99 & 5.69 & 12/110 (10.9\%) & 3.56 \\
\bottomrule
\end{tabular}}
\end{table*}

%% file: multi/multisource_rank_summary.tex
\begin{table*}[t]
\centering
\caption{Relative rank changes after multi-source training.}
\label{tab:multisource_rank_summary}
\resizebox{\linewidth}{!}{ \renewcommand{\arraystretch}{1.0}
\begin{tabular}{lcc|cc|cc|cc}
\toprule
& \multicolumn{2}{c|}{Seen MAE} & \multicolumn{2}{c|}{Seen P@10}
& \multicolumn{2}{c|}{Unseen MAE} & \multicolumn{2}{c}{Unseen P@10} \\
\textbf{Method} & Rank & $\Delta$ & Rank & $\Delta$ & Rank & $\Delta$ & Rank & $\Delta$ \\
\midrule
SimGNN & 13.17 $\to$ 8.33 & +4.83 & 16.08 $\to$ 11.58 & +4.50 & 11.20 $\to$ 10.00 & +1.20 & 16.20 $\to$ 11.60 & +4.60 \\
GMN-emb & 10.67 $\to$ 4.50 & +6.17 & 8.67 $\to$ 4.42 & +4.25 & 2.00 $\to$ 3.40 & -1.40 & 2.20 $\to$ 4.60 & -2.40 \\
GMN-match & 3.83 $\to$ 3.00 & +0.83 & 5.58 $\to$ 4.17 & +1.42 & 1.80 $\to$ 3.80 & -2.00 & 2.40 $\to$ 4.40 & -2.00 \\
GraphSim & 13.17 $\to$ 12.67 & +0.50 & 14.00 $\to$ 13.67 & +0.33 & 13.00 $\to$ 12.00 & +1.00 & 8.80 $\to$ 12.40 & -3.60 \\
H2MN & 13.00 $\to$ 9.00 & +4.00 & 13.00 $\to$ 9.58 & +3.42 & 11.60 $\to$ 10.20 & +1.40 & 14.60 $\to$ 11.20 & +3.40 \\
TaGSim & 14.50 $\to$ 13.17 & +1.33 & 16.67 $\to$ 16.92 & -0.25 & 17.20 $\to$ 14.80 & +2.40 & 19.80 $\to$ 18.00 & +1.80 \\
EGSC & 6.33 $\to$ 17.67 & -11.33 & 8.17 $\to$ 15.50 & -7.33 & 12.40 $\to$ 13.80 & -1.40 & 14.40 $\to$ 15.20 & -0.80 \\
ERIC & 6.50 $\to$ 15.67 & -9.17 & 5.50 $\to$ 13.25 & -7.75 & 4.20 $\to$ 10.20 & -6.00 & 6.20 $\to$ 11.60 & -5.40 \\
GREED & 8.67 $\to$ 6.17 & +2.50 & 9.50 $\to$ 8.00 & +1.50 & 11.80 $\to$ 10.00 & +1.80 & 7.40 $\to$ 7.20 & +0.20 \\
Graph2Region & 8.00 $\to$ 10.50 & -2.50 & 7.58 $\to$ 7.67 & -0.08 & 8.00 $\to$ 9.40 & -1.40 & 9.20 $\to$ 5.20 & +4.00 \\
GraSP & 2.67 $\to$ 2.33 & +0.33 & 4.33 $\to$ 3.25 & +1.08 & 8.40 $\to$ 7.20 & +1.20 & 6.00 $\to$ 5.20 & +0.80 \\
GEN & 2.83 $\to$ 2.00 & +0.83 & 2.08 $\to$ 2.08 & +0.00 & 14.40 $\to$ 8.80 & +5.60 & 6.80 $\to$ 4.80 & +2.00 \\
GOTSim & 18.83 $\to$ 18.67 & +0.17 & 18.33 $\to$ 20.00 & -1.67 & 18.00 $\to$ 18.60 & -0.60 & 9.40 $\to$ 19.80 & -10.40 \\
GEDGNN & 16.50 $\to$ 14.50 & +2.00 & 16.50 $\to$ 14.00 & +2.50 & 9.20 $\to$ 12.00 & -2.80 & 14.90 $\to$ 14.20 & +0.70 \\
GraphEdX & 3.17 $\to$ 3.83 & -0.67 & 3.08 $\to$ 3.58 & -0.50 & 6.75 $\to$ 2.20 & +4.55 & 1.75 $\to$ 2.20 & -0.45 \\
GEDIOT & 13.33 $\to$ 20.00 & -6.67 & 13.67 $\to$ 18.67 & -5.00 & 15.20 $\to$ 19.40 & -4.20 & 12.10 $\to$ 16.20 & -4.10 \\
NOAH* & 11.50 $\to$ 11.33 & +0.17 & 14.92 $\to$ 15.42 & -0.50 & 6.60 $\to$ 8.20 & -1.60 & 13.60 $\to$ 10.90 & +2.70 \\
GENN-A* & 12.33 $\to$ 9.33 & +3.00 & 13.67 $\to$ 9.58 & +4.08 & 13.00 $\to$ 9.80 & +3.20 & 17.60 $\to$ 10.80 & +6.80 \\
MATA* & 14.50 $\to$ 12.33 & +2.17 & 12.33 $\to$ 12.67 & -0.33 & 9.20 $\to$ 13.00 & -3.80 & 16.80 $\to$ 17.50 & -0.70 \\
GELATO & 16.50 $\to$ 15.00 & +1.50 & 6.33 $\to$ 6.00 & +0.33 & 13.40 $\to$ 13.20 & +0.20 & 6.20 $\to$ 7.00 & -0.80 \\
\bottomrule
\end{tabular}}
\end{table*}

%% file: 10appendix.tex
\section*{Appendix}
\startcontents[appendix]

\printcontents[appendix]{}{1}{}
\section{Additional Details about \nm}
\subsection{Benchmark Preprocessing}
\label{app:benchmark}
As described in the main paper, the benchmark uses undirected graphs with categorical node labels and unlabeled edges. We therefore standardize the graph representation across collections before exact GED computation. All graphs are converted to undirected graphs, with self-loops removed and node identifiers relabeled to consecutive integers. Node labels are represented as scalar categorical values when available. For IMDB-Binary, no node labels are available, and GED is therefore computed on unlabeled graphs. For ogbg-code2, whose original node features are multi-dimensional, we use the first feature dimension as the categorical node label and discard edge attributes.

\subsection{Construction of the Resource Landscape}
\label{app:resource_landscape}
We construct the resource landscape to compare the scale and exact supervision available in existing GED resources. Each point represents one graph collection. The horizontal axis reports the median number of nodes among graphs covered by exact GED annotations, characterizing the graph scale of the collection.

Let $P$ denote the number of exact graph pairs in a collection and $U$ the number of unique graphs appearing in at least one exact pair. We measure the amount of exact pairwise supervision per covered graph as
\begin{equation}
D_{\mathrm{pair}}
=
\frac{2P}{U}.
\end{equation}
Intuitively, $D_{\mathrm{pair}}$ is the average number of exact-pair incidences associated with each covered graph. We use this quantity for the vertical axis and display it on a logarithmic scale because it varies substantially across resources.

The bubble size represents $U$, i.e., the number of graphs covered by exact GED supervision. The resulting landscape therefore summarizes  the graph scale, the amount of exact pairwise supervision per covered graph, and the number of graphs receiving exact supervision of each collection.

\subsection{Comparison with Other Resources}
\label{app:resource_comparison}
Existing GED resources differ not only in scale, but also in how exact supervision is constructed and released. Table~\ref{tab:resource_comparison} compares several commonly used resources along five aspects that directly affect benchmark usability: whether graphs are connected, whether isomorphic duplicates are removed, whether optimal node alignments and solver runtimes are released, the timeout used for exact computation, and the underlying GED solver. These differences are important when exact GED pairs are reused for training, evaluation, or further analysis.

\begin{table}[t]
\centering
\caption{Comparison of GED benchmark resources. 
A checkmark indicates that the property is consistently satisfied by the released benchmark used in our comparison. ``n.r.'' indicates that a fixed timeout is not explicitly reported.}
\label{tab:resource_comparison}
\resizebox{0.82\linewidth}{!}{
\renewcommand{\arraystretch}{1.0}
\begin{tabular}{lcccccc}
\toprule
\textbf{Resource} 
& \textbf{Connected} 
& \textbf{Remove Iso.} 
& \textbf{Alignment} 
& \textbf{Runtime} 
& \textbf{Timeout}
& \textbf{GED Solver} \\
\midrule
Classic
& \checkmark
& $\times$
& $\times$
& $\times$
& -- 
& A* / Approx. \\

GraphEdX
& $\times$
& $\times$
& $\times$
& \checkmark
& n.r.
& MIP-F2 \\

Gelato
& $\times$
& $\times$
& \checkmark
& $\times$
& n.r.
& MIP-F2 \\

GRAIL
& $\times$
& $\times$
& $\times$
& $\times$
& 600 s
& MIP-F2 \\

Ours
& \checkmark
& \checkmark
& \checkmark
& \checkmark
& 300 s
& AStar-LSa \\
\bottomrule
\end{tabular}}
\end{table}

\section{Analysis Measures and Computation Details}
\label{app:analysis}

This section provides the computation details for the two structural analyses in Figures~\ref{fig:pair_correlation} and~\ref{fig:collection_shift}. The following subsections define the pair-level quantities used in these analyses.

\subsection{Pair-Level Structural Descriptors}
\label{app:pair_descriptors}

For a graph pair $(G_i,G_j)$, we use four count-based descriptors to characterize its size and the difference between the two graphs. The mean numbers of nodes and edges are defined as
\begin{equation}
\operatorname{mean}|V|(G_i,G_j)
=
\frac{|V_i|+|V_j|}{2},
\qquad
\operatorname{mean}|E|(G_i,G_j)
=
\frac{|E_i|+|E_j|}{2}.
\end{equation}
The absolute differences in node and edge counts are computed as
\begin{equation}
|\Delta V|(G_i,G_j)
=
\left||V_i|-|V_j|\right|,
\qquad
|\Delta E|(G_i,G_j)
=
\left||E_i|-|E_j|\right|.
\end{equation}
All descriptors are computed for every exact graph pair in the benchmark.
\subsection{Graphlet Distance}
\label{app:graphlet_distance}

We use graphlets to characterize differences in local topology. Specifically, we consider connected induced graphlets with three or four nodes. The three-node graphlets include a three-node path and a triangle, while the four-node graphlets include a four-node path, a three-star, a four-cycle, a tailed triangle, a diamond, and a four-node clique.

Let $\mathbf{c}_k(G)$ denote the graphlet-count vector of order $k\in\{3,4\}$. To represent the relative composition of graphlet types within each order, we normalize the counts as
\begin{equation}
\mathbf{f}_k(G)
=
\frac{\mathbf{c}_k(G)}
{\sum_j c_{k,j}(G)}.
\end{equation}
If a graph contains no connected graphlet of order $k$, we set $\mathbf{f}_k(G)$ to the zero vector. For a graph pair $(G_i,G_j)$, we compare their graphlet compositions using the total variation distance, as follows
\begin{equation}
\operatorname{TV}
\left(
\mathbf{f}_k(G_i),\mathbf{f}_k(G_j)
\right)
=
\frac{1}{2}
\left\|
\mathbf{f}_k(G_i)-\mathbf{f}_k(G_j)
\right\|_1.
\end{equation}
We then average the distances for the two graphlet orders,
\begin{equation}
d_{\mathrm{graphlet}}(G_i,G_j)
=
\frac{1}{2}
\left[
\operatorname{TV}
\left(
\mathbf{f}_3(G_i),\mathbf{f}_3(G_j)
\right)
+
\operatorname{TV}
\left(
\mathbf{f}_4(G_i),\mathbf{f}_4(G_j)
\right)
\right].
\end{equation}
The resulting graphlet distance gives equal weight to three- and four-node patterns and summarizes the difference in local structural composition between the two graphs.

\subsection{Spectral Distance}
\label{app:spectral_distance}

We use spectral distance to characterize differences in global connectivity. For a graph $G$ with adjacency matrix $A$ and degree matrix $D$, we first construct the normalized graph Laplacian
\begin{equation}
L_{\mathrm{norm}}
=
I-D^{-1/2}AD^{-1/2}.
\end{equation}
Let $\lambda_1,\ldots,\lambda_{|V|}$ denote its eigenvalues. We represent the spectrum of $G$ as the empirical distribution
\begin{equation}
\mu_G
=
\frac{1}{|V|}
\sum_{i=1}^{|V|}
\delta_{\lambda_i},
\end{equation}
where each eigenvalue has equal mass. This representation retains the full spectrum and allows graphs with different numbers of nodes to be compared without padding or truncation. For a graph pair $(G_i,G_j)$, we compare their spectral distributions using the Wasserstein-1 distance, which measures how far the eigenvalue mass of one spectral distribution must move along the eigenvalue axis to match the other, as follows
\begin{equation}
W_1(\mu_{G_i},\mu_{G_j})
=
\int_{-\infty}^{\infty}
\left|
F_{G_i}(x)-F_{G_j}(x)
\right|\,dx,
\end{equation}
where $F_{G_i}$ and $F_{G_j}$ are the corresponding cumulative distribution functions.  Because the eigenvalues of the normalized graph Laplacian lie in $[0,2]$, we normalize the distance as
\begin{equation}
d_{\mathrm{spectral}}(G_i,G_j)
=
\frac{1}{2}
W_1(\mu_{G_i},\mu_{G_j}).
\end{equation}
The resulting spectral distance lies in $[0,1]$, with larger values indicating greater differences in global connectivity.

\subsection{Structural Factors Associated with GED}
\label{app:within_collection_structure}
We measure the association between each structural factor and GED separately within each collection using Spearman's rank correlation. Intuitively, Spearman correlation asks whether graph pairs that rank higher on a structural factor also tend to rank higher in GED.

For collection $c$ with $N_c$ exact graph pairs, let $x_{k,n}^{(c)}$ denote the value of structural factor $k$ for pair $n$, and let $y_n^{(c)}$ denote its exact GED. We compute
\begin{equation}
\rho_k^{(c)}
=
\operatorname{Spearman}
\left(
\{x_{k,n}^{(c)}\}_{n=1}^{N_c},
\{y_n^{(c)}\}_{n=1}^{N_c}
\right).
\end{equation}

\begin{figure}[h]
    \centering
    \includegraphics[width=\linewidth]{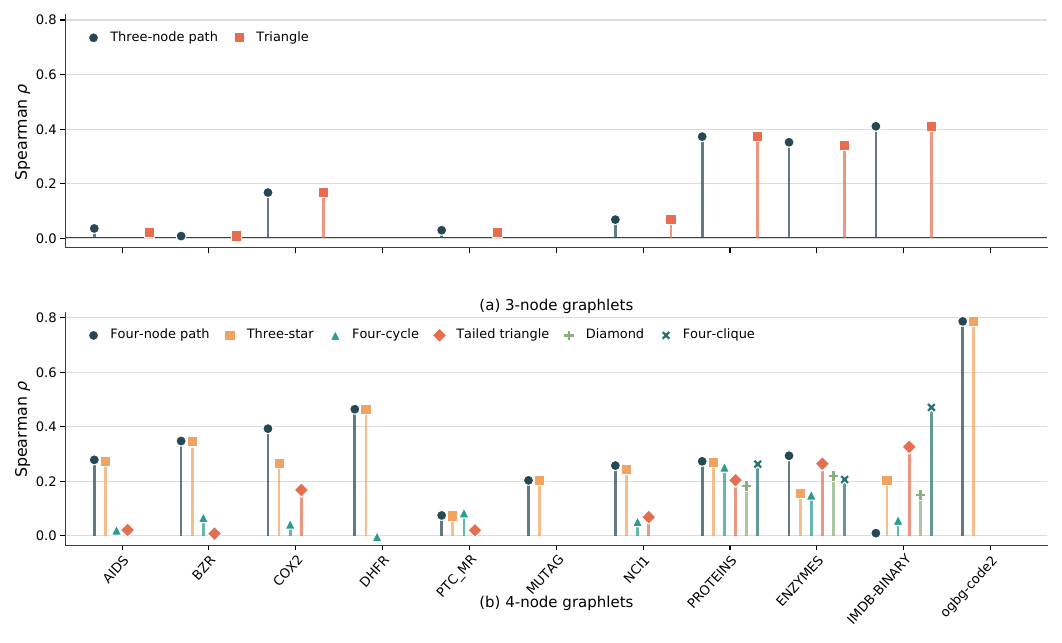}
    \caption{Spearman correlations between exact GED and the frequency difference of each graphlet type.}
    \label{fig:graphlet_type_correlation}
\end{figure}

For clarity in the main-text comparison, Figure~\ref{fig:pair_correlation} represents local structural differences using the aggregate graphlet distance defined above. In this appendix, we further examine these graphlet types individually to determine whether their associations with GED differ across collections. For graphlet type $t$, we define the pairwise frequency difference as
\begin{equation}
\Delta f_t(G_i,G_j)
=
\left|
f_t(G_i)-f_t(G_j)
\right|,
\end{equation}
where $f_t(G)$ is the normalized frequency of type $t$ within its graphlet order. We then compute the Spearman correlation between $\Delta f_t$ and exact GED within each collection.

Figure~\ref{fig:graphlet_type_correlation} shows that the graphlet types most strongly associated with GED differ across collections. In ogbg-code2, the four-node path and three-star both reach a correlation of $0.786$. In IMDB-Binary, the four-node path is nearly uncorrelated with GED at $0.010$, while the four-node clique, triangle, and three-node path reach $0.471$, $0.410$, and $0.410$, respectively. The same local structural pattern can therefore carry very different GED signals across collections.

\subsection{Collection Similarity under Different Structural Views}
\label{app:collection_shift}
We compare graph collections by the distributions of their structural descriptors. As described in the main paper, we consider four structural views: graph-level properties, pair-level count properties, graphlet distance, and spectral distance. Each view characterizes a different aspect of the structural distribution of a collection.

For the graph-level view, we use the number of nodes, number of edges, graph density, and average degree of each graph. For the pair-level count view, we use the four descriptors defined in Section~\ref{app:pair_descriptors} together with the node-count ratio
\begin{equation}
R_V(G_i,G_j)
=
\frac{\max(|V_i|,|V_j|)}
     {\min(|V_i|,|V_j|)}.
\end{equation}
The graphlet and spectral views use the distributions of $d_{\mathrm{graphlet}}$ and $d_{\mathrm{spectral}}$, respectively, over all exact graph pairs in each collection.

For each descriptor $s$, let $P_s^{(a)}$ and $P_s^{(b)}$ denote its empirical distributions in collections $a$ and $b$. We measure their difference using the Wasserstein-1 distance,
\begin{equation}
D_s(a,b)
=
W_1
\left(
P_s^{(a)},
P_s^{(b)}
\right).
\end{equation}
Because different descriptors have different numerical ranges, we normalize each descriptor by its largest distance among all collection pairs,
\begin{equation}
\widetilde{D}_s(a,b)
=
\frac{D_s(a,b)}
{\max_{u<v} D_s(u,v)}.
\end{equation}
The resulting value lies in $[0,1]$ and represents the relative distribution shift for descriptor $s$.

For a structural view containing multiple descriptors, we average their normalized distances. Let $\mathcal{S}_v$ denote the set of descriptors in view $v$. The distance between collections $a$ and $b$ under this view is
\begin{equation}
D_v(a,b)
=
\frac{1}{|\mathcal{S}_v|}
\sum_{s\in\mathcal{S}_v}
\widetilde{D}_s(a,b).
\end{equation}
For the graphlet and spectral views, each view contains a single descriptor, so the corresponding normalized Wasserstein distance is used directly.

\section{Experimental Setup Details}
\label{app:experimental_setup}

\subsection{Graph Splits and Pair Construction}
\label{app:data_splits}

We split graphs before constructing graph pairs so that the same graph does not appear across training, validation, and test splits. For each collection, $20\%$ of the graphs are first held out from training. Among the held-out graphs, $20\%$ are assigned to validation and the remainder to test. Training pairs are then constructed from training graphs only, while validation pairs are formed between validation and training graphs.

For evaluation, we consider three test views according to which graph splits form a pair. \emph{All} contains all exact pairs with at least one test graph, including test--training, test--validation, and test--test pairs. \emph{D--Q} contains test--training and test--validation pairs, where the test graph acts as an unseen query and the other graph comes from the data available before testing. \emph{Q--Q} contains only test--test pairs, where both graphs are unseen during training. The main paper reports \emph{All}, while \emph{D--Q} and \emph{Q--Q} provide additional views of generalization under different pair compositions.

\subsection{Run Aggregation and Ranking}
\label{app:aggregation}

Our evaluation contains a large number of training and evaluation runs across collections and source--target combinations. Table~\ref{tab:experimental_scale} summarizes the overall experimental scale. Across the 20 methods, the study involves more than 16,000 training or evaluation runs. Reporting each collection and run separately would make the overall comparison difficult to interpret. We therefore first compute performance at the collection or source--target task level and then summarize these results across tasks. This also gives each collection equal weight and prevents collections with substantially more exact graph pairs from dominating the aggregate results.

\begin{table}[t]
\centering
\caption{Scale of the experimental evaluation.}
\label{tab:experimental_scale}
\resizebox{0.82\linewidth}{!}{
\renewcommand{\arraystretch}{1.0}
\begin{tabular}{lrrl}
\toprule
\textbf{Setting}
& \textbf{Runs / Method}
& \textbf{Total}
& \textbf{Configuration} \\
\midrule
In-collection
& 55
& 1,100
& 11 collections $\times$ 5 runs \\

Zero-shot transfer
& 550
& 11,000
& 11 sources $\times$ 10 targets $\times$ 5 runs \\

Multi-source pretraining
& 1
& 20
& Fixed multi-source pool \\

Limited-target fine-tuning
& 45
& 900
& 5 targets $\times$ 3 budgets $\times$ 3 runs \\

Few-shot / Scratch
& 165
& 3,300
& 11 targets $\times$ 3 budgets $\times$ 5 runs \\
\midrule
\textbf{Total}
& \textbf{816}
& \textbf{16,320}
& \\
\bottomrule
\end{tabular}}
\end{table}

For the single-source experiments, including both in-collection evaluation and zero-shot transfer, we use five runs with $(\text{split seed}, \text{training seed})$ equal to $(1729,0)$, $(1729,1)$, $(1729,2)$, $(3407,0)$, and $(9679,0)$. Several methods exhibit heavy-tailed performance across runs, where a small number of runs can produce substantially larger errors than the rest. We therefore use the median rather than the mean when aggregating metric values.

Because split 1729 contains three training seeds while the other two splits contain one each, directly taking the median over all five runs would give greater weight to split 1729. We therefore use a split-balanced median when reporting MAE and Precision@10. For each graph split, we first take the median across its available training seeds and then take the median across the three split-level values. Each graph split therefore contributes equally to the reported metric value.

Ranking is performed separately from metric aggregation. For each source--target task and run, methods are ranked by MAE in ascending order and Precision@10 in descending order, with rank 1 indicating the best result. Runs with invalid or missing results are excluded from the corresponding comparison. The valid run-level ranks are first averaged to obtain a rank for each source--target task, and these task-level ranks are then averaged over the corresponding in-collection or cross-collection tasks.

\section{Single-Source Generalization Details}
\label{app:single_source}
The main paper reports the overall single-source generalization results and shows that strong in-collection performance does not generally carry over to unseen collections. This section provides the additional details behind these results. We first describe the approximation baselines used for comparison and the construction of the relative-MAE results. We then report the ranking statistics and full source--target results, followed by separate results under the $D$--$Q$ and $Q$--$Q$ test views.

\subsection{Approximation Baselines}
\label{app:approximation_baselines}
We compare the learning-based methods against four GED approximation baselines. \emph{Hungarian} formulates graph matching as a bipartite assignment problem and derives an approximate GED from the resulting matching. \emph{VJ} follows a similar matching-based formulation but uses the Volgenant--Jonker assignment solver. \emph{Beam} searches over candidate node matchings using beam search. \emph{FGWAlign} \citep{fgwalign} aligns two graphs using Fused Gromov--Wasserstein optimal transport. We report the performance of all four approximation methods on the 11 graph collections. Because no single approximation method performs best on every graph pair, we additionally construct a pair-wise oracle by selecting, for each graph pair, the approximation result with the smallest absolute error to the exact GED. MAE and Precision@10 are then computed from these pair-wise oracle predictions. The performance of the four individual methods and the resulting oracle is reported in Tables~\ref{tab:approximation_mae} and~\ref{tab:approximation_pat10}, respectively.

\input{appro/mae}
\input{appro/p10}

\subsection{Construction of Relative-MAE Results}
\label{app:single_source_relative_mae}
To compare neural GED methods with approximation baselines across collections, we normalize each neural MAE by the oracle approximation MAE of the target collection. For method $m$ trained on source collection $s$ and evaluated on target collection $t$, we define
\begin{equation}
R_{m}^{s\rightarrow t}
=
\log_2
\frac{
\mathrm{MAE}_{m}^{s\rightarrow t}
}{
\mathrm{MAE}_{\mathrm{oracle}}^{t}
}.
\end{equation}

The oracle MAE for each target collection is taken from Table~\ref{tab:approximation_mae}. The resulting relative-MAE value is centered at $0$. $R=0$ means that the neural method matches the oracle approximation, $R<0$ means that it achieves a lower MAE, and $R>0$ means that it performs worse. We compute this quantity for every valid source--target result and use it to construct the relative-MAE comparisons reported in the main paper.

\subsection{Ranking and Top-$k$ Statistics}
\label{app:single_source_ranking}

To summarize the relative performance of the learning-based methods across source--target pairs, we use the task-level ranks defined in Section~\ref{app:aggregation}. For each method, we separately average its ranks over in-collection tasks, where the source and target collections are the same, and cross-collection tasks, where they differ. Lower values indicate better relative performance. We additionally report the Top-$k$ Rate to show how often a method remains among the best-performing methods across source--target tasks. 
\subsection{Full Results}
\label{app:single_source_full}

\begin{figure}[h]
    \centering
    \includegraphics[width=\linewidth]{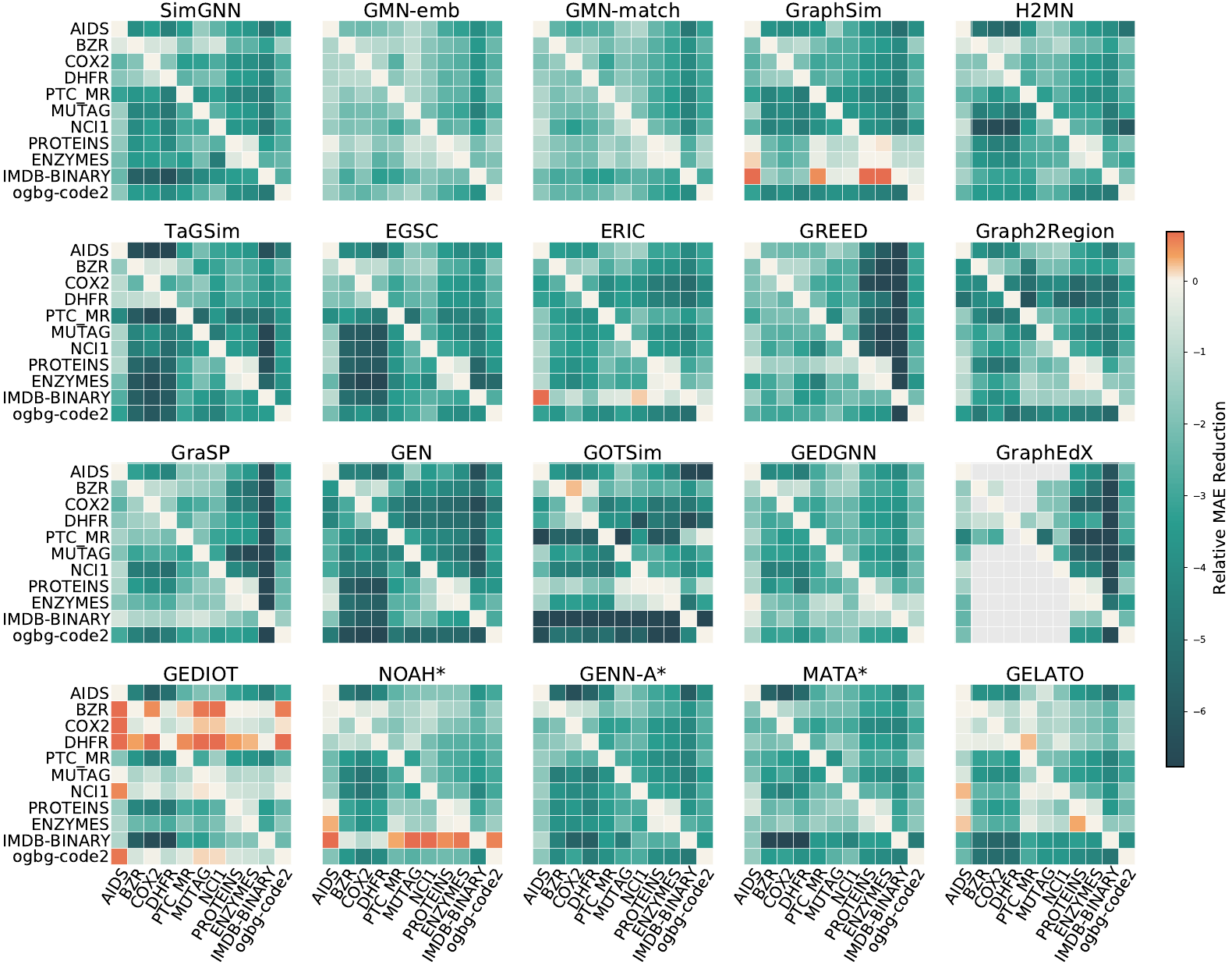} 
    \caption{Source-to-target MAE changes.}
    \label{fig:single_source_mae_heatmaps}
\end{figure}

\begin{figure}[h]
    \centering
    \includegraphics[width=\linewidth]{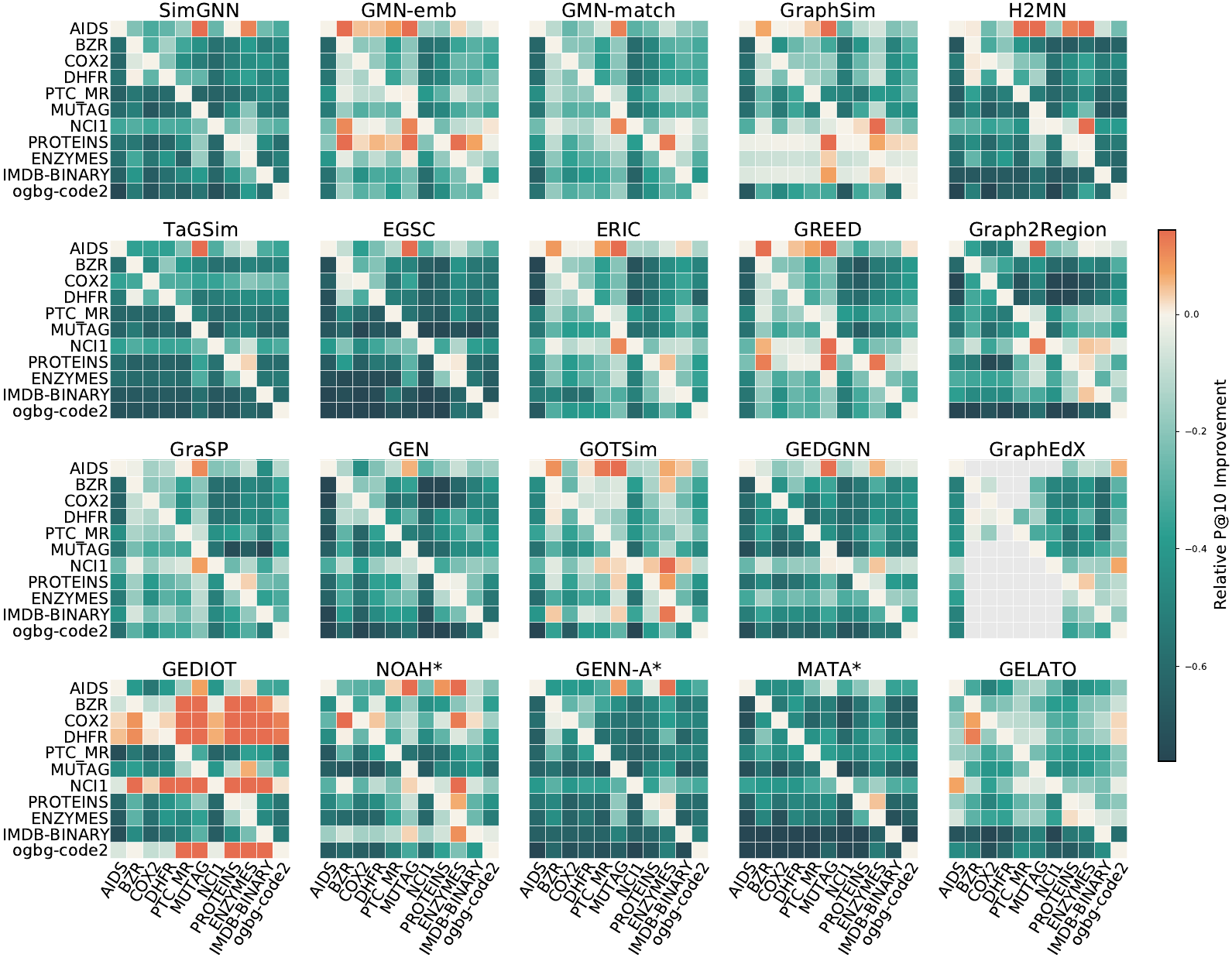} 
    \caption{Source-to-target Precision@10 changes.}
    \label{fig:single_source_pat10_heatmaps}
\end{figure}

To provide a complete view of source--target transfer, Figures~\ref{fig:single_source_mae_heatmaps} and~\ref{fig:single_source_pat10_heatmaps} report the results of all 20 methods under the \emph{All} test view. For MAE, we use the same base-2 log-ratio convention as above, but take the corresponding in-collection result $\mathrm{MAE}^{s\rightarrow s}$ as the reference. The diagonal is therefore zero, while positive values indicate lower MAE after transfer. For Precision@10, we report the difference from the corresponding in-collection result, with negative values indicating a drop in retrieval performance.

Figures~\ref{fig:single_source_mae_heatmaps} and~\ref{fig:single_source_pat10_heatmaps} reveal a clear directional asymmetry in cross-collection transfer. Transferring a model from collection $s$ to collection $t$ can produce a different degradation from transferring the same method from $t$ to $s$. This behavior appears across many methods and under both metrics. For example, AIDS$\rightarrow$IMDB-Binary shows much larger MAE degradation than IMDB-Binary$\rightarrow$AIDS for most methods, while similarly strong directional differences appear for several other collection pairs. Transferability is therefore not solely a property of a pair of collections; it also depends on which collection serves as the source and which serves as the target.

Notably, some entries are unavailable for GraphEdX. Its node-alignment layers are parameterized by a source-dependent \texttt{max\_node\_set\_size}, which fixes the output dimension of the node Sinkhorn feature layers. A source-trained GraphEdX model therefore cannot directly process target graphs whose number of nodes exceeds the maximum size supported by the source model.

\subsection{$D$--$Q$ and $Q$--$Q$ Results}
\label{app:dq_qq}

To examine whether the single-source conclusions depend on the composition of test pairs, we additionally report the ranking statistics under the $D$--$Q$ and $Q$--$Q$ views defined in Section~\ref{app:data_splits}. Tables~\ref{tab:single_source_rank_dq} and~\ref{tab:single_source_rank_qq} report the corresponding results.

The ranking patterns remain highly consistent across the three test-pair views. Compared with the \emph{All} view in Table~\ref{tab:single_source_rank}, the cross-collection MAE ranks change only marginally under both $D$--$Q$ and $Q$--$Q$. The corresponding rank correlations with the \emph{All} view are $0.9996$ and $0.9970$, respectively. Larger differences appear mainly in the in-collection results and Precision@10 under $Q$--$Q$. Overall, the cross-collection MAE ranking is largely insensitive to whether one or both graphs in a test pair are unseen during training.
\input{single/dq}
\input{single/qq}

\section{Multi-Source Generalization Details}
\label{app:multi_source}

This section provides additional details for the multi-source generalization experiments in the main paper. We first describe the source collections and training procedure, then define the reference baselines for seen and unseen targets and the relative performance changes used in the analysis. Finally, we report the full results for all methods.

\subsection{Multi-Source Training}
\label{app:multi_source_training}
We train each method jointly on seven source collections: AIDS, BZR, ENZYMES, MUTAG, NCI1, PTC-MR, and LINUX. The first six are benchmark collections and are therefore also evaluated as seen targets. LINUX, provided in \citep{simgnn}, after removing isomorphic graphs, is used only to increase source diversity during multi-source training and is not included as an evaluation target. The remaining five benchmark collections, PROTEINS, IMDB-Binary, ogbg-code2, COX2, and DHFR, are held out from multi-source training and used as unseen targets.

To prevent collections with more exact pairs from dominating training, each source contributes at most 100 mini-batches per epoch. Validation is performed only on exact pairs from the source collections. We compute the validation MSE separately for each source and use their macro average for model selection, giving each source collection equal weight.

\subsection{Construction of Relative Performance Changes}
\label{app:multi_source_relative}

We compare multi-source training against different single-source references for seen and unseen targets. For a seen target $t$, the reference is the target-specific single-source result obtained by training directly on $t$. For an unseen target, no target-specific model is available, so we use the median performance over all valid single-source zero-shot transfers $s\rightarrow t$ with $s\neq t$. The former measures whether multi-source training can match direct supervision on a seen collection, while the latter measures whether it improves over typical single-source transfer to an unseen collection. Following the relative-performance convention used in Section~\ref{app:single_source_relative_mae}, we report changes on a base-2 logarithmic scale, with the direction adjusted so that positive values consistently indicate improvement from multi-source training. A value of zero indicates no change, while negative values indicate worse performance.

\section{Limited-Target Adaptation Details}
\label{app:adaptation}

This section provides additional details for the limited-target adaptation experiments in the main paper.

\subsection{Adaptation Setup}
\label{app:adaptation_setup}

We evaluate adaptation on the five collections held out from multi-source training: PROTEINS, IMDB-Binary, ogbg-code2, COX2, and DHFR. For each target collection, we construct supervision budgets of 100, 500, and 2,000 exact graph pairs. The three budgets are nested, so that the pairs used at a smaller budget are always included in the larger one.

For each budget, we compare two training settings using exactly the same target pairs. \emph{Pretrained} initializes the model from the multi-source checkpoint and fine-tunes it on the target supervision, while \emph{Scratch} trains the same model using only the available target pairs. This isolates the effect of multi-source pretraining from the amount of target supervision.

\subsection{Construction of Relative MAE Reduction}
\label{app:adaptation_relative}

We measure the benefit of multi-source pretraining by comparing the MAE obtained after fine-tuning with that of training from scratch under the same target-supervision budget. Relative MAE reduction is reported as the percentage decrease from the scratch result,
\begin{equation}
\mathrm{Reduction}
=
\frac{
\mathrm{MAE}_{\mathrm{scratch}}
-
\mathrm{MAE}_{\mathrm{pretrained}}
}{
\mathrm{MAE}_{\mathrm{scratch}}
}
\times 100\%.
\end{equation}
Positive values indicate that multi-source pretraining improves adaptation, while negative values indicate that training from scratch performs better.

\begin{figure}[h]
    \centering
    \includegraphics[width=\linewidth]{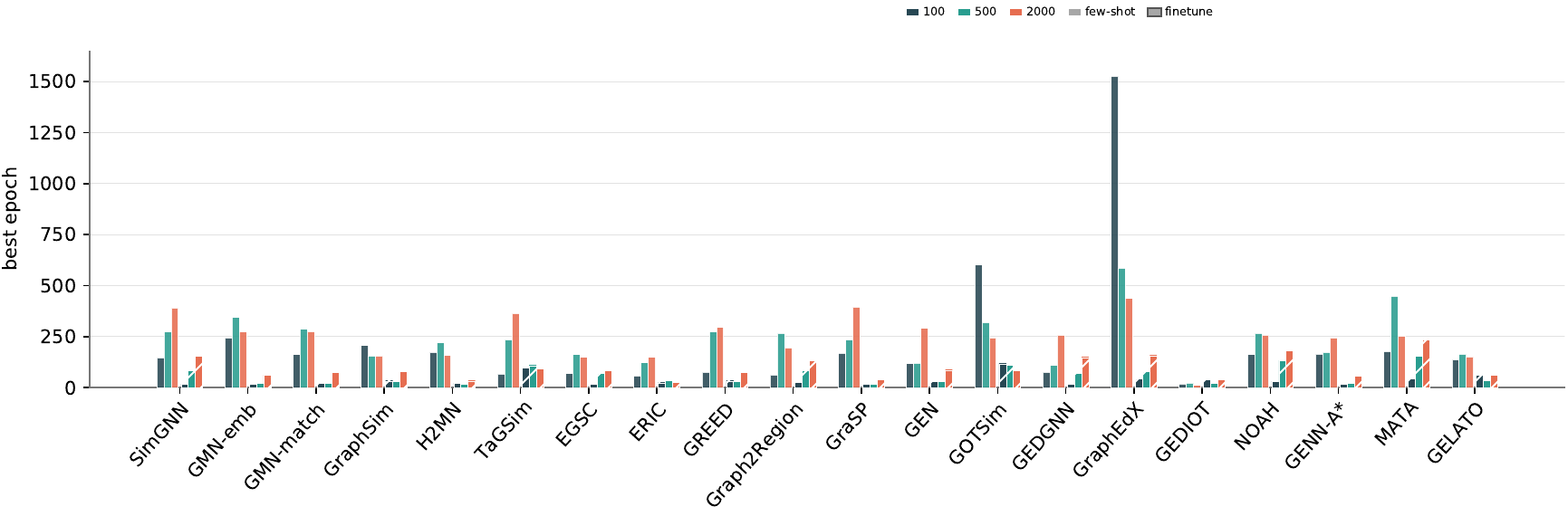} 
    \caption{Best validation epochs for limited-target adaptationand training from scratch.}
    \label{fig:adaptation_best_epoch}
\end{figure}

We further compare the best validation epochs reached with and without multi-source pretraining to examine adaptation speed. 
Figure~\ref{fig:adaptation_best_epoch} shows that fine-tuning reaches its best checkpoint earlier in 57 of the 60 method--budget settings. Across the 20 methods, the median best epoch decreases from $153.5$ to $28$ with 100 target pairs, from $234$ to $34$ with 500 pairs, and from $254.5$ to $79$ with 2,000 pairs. The median reductions across methods are $78.6\%$, $73.8\%$, and $65.4\%$, respectively. Multi-source pretraining therefore improves not only adaptation accuracy but also convergence speed. This advantage becomes smaller as more target supervision becomes available.

\stopcontents[appendix]

%% file: appro/mae.tex
\begin{table*}[t]
\centering
\caption{MAE of approximation baselines across graph collections.}
\label{tab:approximation_mae}
\resizebox{0.8\linewidth}{!}{ \renewcommand{\arraystretch}{1.0}
\begin{tabular}{l|cccc|c}
\toprule
Collection & Hungarian & VJ & Beam & FGWAlign & Oracle Approx. \\
\midrule
AIDS & 9.1748 & 9.7393 & 3.4365 & 1.6830 & 1.2106 \\
BZR & 30.4388 & 41.4946 & 14.4075 & 9.1885 & 5.8971 \\
COX2 & 40.2620 & 56.0973 & 15.7913 & 10.4754 & 6.4562 \\
DHFR & 32.8957 & 51.0385 & 12.6850 & 13.5718 & 6.7223 \\
PTC\_MR & 8.3955 & 8.9508 & 4.9713 & 3.8320 & 2.3268 \\
MUTAG & 9.0095 & 9.3844 & 6.5564 & 1.9485 & 1.5965 \\
NCI1 & 20.0207 & 19.5561 & 10.3495 & 5.1388 & 4.4886 \\
PROTEINS & 17.9815 & 17.7238 & 13.3342 & 2.3360 & 2.1639 \\
ENZYMES & 20.8244 & 20.5983 & 16.4363 & 2.4995 & 2.3688 \\
IMDB-BINARY & 37.8413 & 37.8413 & 24.4503 & 1.4924 & 1.2358 \\
ogbg-code2 & 6.4123 & 5.9527 & 4.1000 & 6.7472 & 2.1805 \\
\bottomrule
\end{tabular}}
\end{table*}

%% file: appro/p10.tex
\begin{table*}[t]
\centering
\caption{P@10 of approximation baselines across graph collections.}
\label{tab:approximation_pat10}
\resizebox{0.8\linewidth}{!}{ \renewcommand{\arraystretch}{1.0}
\begin{tabular}{l|cccc|c}
\toprule
Collection & Hungarian & VJ & Beam & FGWAlign & Oracle Approx. \\
\midrule
AIDS & 0.1811 & 0.1832 & 0.6521 & 0.6336 & 0.8151 \\
BZR & 0.4250 & 0.4266 & 0.6469 & 0.6312 & 0.8219 \\
COX2 & 0.4560 & 0.4187 & 0.7400 & 0.6053 & 0.8227 \\
DHFR & 0.4617 & 0.4713 & 0.6348 & 0.6061 & 0.7730 \\
PTC\_MR & 0.3904 & 0.3750 & 0.6231 & 0.5577 & 0.7250 \\
MUTAG & 0.5679 & 0.5714 & 0.6643 & 0.7750 & 0.8536 \\
NCI1 & 0.1773 & 0.1713 & 0.3352 & 0.4257 & 0.4930 \\
PROTEINS & 0.2724 & 0.2897 & 0.4121 & 0.6431 & 0.6793 \\
ENZYMES & 0.4105 & 0.4000 & 0.4421 & 0.7737 & 0.7842 \\
IMDB-BINARY & 0.2333 & 0.2333 & 0.5778 & 0.8800 & 0.9089 \\
ogbg-code2 & 0.6129 & 0.6032 & 0.8129 & 0.7194 & 0.8935 \\
\bottomrule
\end{tabular}}
\end{table*}

%% file: single/dq.tex
\begin{table*}[t]
\centering
\caption{Single-Source Generalization rank summary for the D-Q (test versus train/validation) test-pair view. GEDRanker is excluded before run-level ranking.}
\label{tab:single_source_rank_dq}
\label{tab:approximation_mae}
\resizebox{\linewidth}{!}{ \renewcommand{\arraystretch}{1.0}
\begin{tabular}{lccc|ccc|c}
\toprule
\multirow{2}{*}{Method} & \multicolumn{3}{c|}{In-collection} & \multicolumn{3}{c|}{Cross-collection} & \multirow{2}{*}{$\Delta$ Rank} \\
& MAE Rank & P@10 Rank & Top-5 Rate & MAE Rank & P@10 Rank & Top-5 Rate & \\
\midrule
SimGNN & 12.96 & 14.35 & 0/11 (0.0\%) & 11.49 & 14.54 & 0/110 (0.0\%) & 1.47 \\
GMN-emb & 10.24 & 8.96 & 1/11 (9.1\%) & 4.01 & 3.96 & 84/110 (76.4\%) & 6.23 \\
GMN-match & 3.91 & 5.94 & 8/11 (72.7\%) & 2.95 & 4.36 & 98/110 (89.1\%) & 0.95 \\
GraphSim & 13.24 & 13.58 & 0/11 (0.0\%) & 12.26 & 10.92 & 2/110 (1.8\%) & 0.97 \\
H2MN & 12.29 & 12.48 & 0/11 (0.0\%) & 11.00 & 13.39 & 4/110 (3.6\%) & 1.29 \\
TaGSim & 14.09 & 16.41 & 0/11 (0.0\%) & 13.83 & 17.86 & 2/110 (1.8\%) & 0.26 \\
EGSC & 7.85 & 8.06 & 2/11 (18.2\%) & 11.05 & 13.28 & 3/110 (2.7\%) & -3.20 \\
ERIC & 6.89 & 5.61 & 4/11 (36.4\%) & 8.14 & 5.71 & 23/110 (20.9\%) & -1.25 \\
GREED & 9.80 & 9.85 & 1/11 (9.1\%) & 9.89 & 6.53 & 32/110 (29.1\%) & -0.09 \\
Graph2Region & 6.56 & 6.81 & 6/11 (54.5\%) & 10.30 & 8.92 & 19/110 (17.3\%) & -3.73 \\
GraSP & 4.96 & 5.89 & 7/11 (63.6\%) & 7.92 & 5.60 & 45/110 (40.9\%) & -2.96 \\
GEN & 3.22 & 3.45 & 10/11 (90.9\%) & 12.67 & 7.54 & 11/110 (10.0\%) & -9.45 \\
GOTSim & 17.78 & 16.76 & 0/11 (0.0\%) & 15.00 & 10.89 & 0/110 (0.0\%) & 2.78 \\
GEDGNN & 15.80 & 15.62 & 0/11 (0.0\%) & 11.18 & 14.83 & 0/110 (0.0\%) & 4.62 \\
GraphEdX & 3.71 & 4.15 & 9/11 (81.8\%) & 9.03 & 3.66 & 23/65 (35.4\%) & -5.32 \\
GEDIOT & 13.80 & 13.60 & 0/11 (0.0\%) & 13.34 & 11.48 & 3/110 (2.7\%) & 0.46 \\
NOAH$^*$ & 13.22 & 15.96 & 0/11 (0.0\%) & 7.25 & 13.03 & 34/110 (30.9\%) & 5.96 \\
GENN-A$^*$ & 11.96 & 13.61 & 0/11 (0.0\%) & 11.65 & 14.90 & 1/110 (0.9\%) & 0.31 \\
MATA$^*$ & 12.93 & 12.01 & 0/11 (0.0\%) & 11.53 & 16.14 & 3/110 (2.7\%) & 1.40 \\
GELATO & 14.78 & 6.88 & 1/11 (9.1\%) & 11.00 & 5.77 & 12/110 (10.9\%) & 3.78 \\
\bottomrule
\end{tabular}}
\end{table*}

%% file: single/qq.tex
\begin{table*}[h]
\centering
\caption{Single-Source Generalization rank summary for the Q-Q (test versus test) test-pair view. GEDRanker is excluded before run-level ranking.}
\label{tab:single_source_rank_qq}
\label{tab:approximation_mae}
\resizebox{\linewidth}{!}{ \renewcommand{\arraystretch}{1.0}
\begin{tabular}{lccc|ccc|c}
\toprule
\multirow{2}{*}{Method} & \multicolumn{3}{c|}{In-collection} & \multicolumn{3}{c|}{Cross-collection} & \multirow{2}{*}{$\Delta$ Rank} \\
& MAE Rank & P@10 Rank & Top-5 Rate & MAE Rank & P@10 Rank & Top-5 Rate & \\
\midrule
SimGNN & 12.31 & 11.87 & 0/11 (0.0\%) & 11.58 & 13.33 & 0/110 (0.0\%) & 0.73 \\
GMN-emb & 11.02 & 11.22 & 0/11 (0.0\%) & 4.07 & 4.97 & 81/110 (73.6\%) & 6.94 \\
GMN-match & 4.75 & 5.59 & 7/11 (63.6\%) & 2.96 & 4.93 & 97/110 (88.2\%) & 1.79 \\
GraphSim & 13.53 & 14.13 & 0/11 (0.0\%) & 12.34 & 10.23 & 3/110 (2.7\%) & 1.19 \\
H2MN & 11.95 & 12.48 & 0/11 (0.0\%) & 11.07 & 12.20 & 3/110 (2.7\%) & 0.87 \\
TaGSim & 13.38 & 13.98 & 0/11 (0.0\%) & 13.76 & 17.03 & 2/110 (1.8\%) & -0.37 \\
EGSC & 7.31 & 6.47 & 2/11 (18.2\%) & 10.97 & 12.42 & 3/110 (2.7\%) & -3.66 \\
ERIC & 7.02 & 6.76 & 5/11 (45.5\%) & 8.15 & 6.96 & 24/110 (21.8\%) & -1.13 \\
GREED & 10.56 & 10.35 & 0/11 (0.0\%) & 9.90 & 6.99 & 31/110 (28.2\%) & 0.66 \\
Graph2Region & 7.75 & 7.96 & 4/11 (36.4\%) & 10.23 & 10.32 & 19/110 (17.3\%) & -2.49 \\
GraSP & 5.51 & 6.81 & 7/11 (63.6\%) & 7.95 & 5.69 & 43/110 (39.1\%) & -2.44 \\
GEN & 2.51 & 3.00 & 9/11 (81.8\%) & 12.66 & 7.61 & 10/110 (9.1\%) & -10.15 \\
GOTSim & 17.58 & 17.26 & 0/11 (0.0\%) & 15.03 & 10.07 & 0/110 (0.0\%) & 2.56 \\
GEDGNN & 15.35 & 13.60 & 0/11 (0.0\%) & 11.18 & 14.79 & 0/110 (0.0\%) & 4.16 \\
GraphEdX & 2.82 & 3.87 & 10/11 (90.9\%) & 9.10 & 5.59 & 23/65 (35.4\%) & -6.28 \\
GEDIOT & 14.65 & 14.33 & 0/11 (0.0\%) & 13.34 & 11.40 & 3/110 (2.7\%) & 1.32 \\
NOAH$^*$ & 12.51 & 13.20 & 1/11 (9.1\%) & 7.28 & 10.99 & 32/110 (29.1\%) & 5.23 \\
GENN-A$^*$ & 11.05 & 10.98 & 0/11 (0.0\%) & 11.71 & 14.25 & 1/110 (0.9\%) & -0.66 \\
MATA$^*$ & 15.07 & 13.93 & 0/11 (0.0\%) & 11.43 & 15.55 & 3/110 (2.7\%) & 3.64 \\
GELATO & 13.38 & 12.19 & 1/11 (9.1\%) & 10.84 & 8.78 & 12/110 (10.9\%) & 2.55 \\
\bottomrule
\end{tabular}}
\end{table*}